\documentclass[journal]{IEEEtran}

\usepackage{times}  
\usepackage{helvet}  
\usepackage{courier}  
\usepackage{url}  
\usepackage{graphicx}  
\usepackage{mathptmx}
\usepackage{amsmath}
\usepackage{amsfonts}
\usepackage{bm}
\usepackage{array}
\begin{document}

\title{Holistic Multi-modal Memory Network  \\for Movie Question Answering}
%
%
%

\author{Anran Wang,
 Anh Tuan Luu, Chuan-Sheng Foo, Hongyuan Zhu, Yi Tay, Vijay Chandrasekhar\\

\thanks{Anran Wang, Anh Tuan Luu, Chuan-Sheng Foo, Hongyuan Zhu, and Vijay Chandrasekhar are with Institute for Infocomm Research, A*STAR, Singapore 138632 (e-mail: wang\_anran@i2r.a-star.edu.sg; at.luu@i2r.a-star.edu.sg; foo\_chuan\_sheng@i2r.a-star.edu.sg; zhuh@i2r.a-star.edu.sg; vijay@i2r.a-star.edu.sg).
}
\thanks{Yi Tay is with the School of Computer Science and Engineering, Nanyang Technological University, Singapore 639798 (e-mail: YTAY017@e.ntu.edu.sg).
}
}

%
%

\markboth{Journal of \LaTeX\ Class Files,~Vol.~14, No.~8, August~2015}%
{Shell \MakeLowercase{\textit{et al.}}: Bare Demo of IEEEtran.cls for IEEE Journals}
%



\maketitle

\begin{abstract}

Answering questions according to multi-modal context is a challenging problem as it requires a deep integration of different data sources. Existing approaches only employ partial interactions among data sources in one attention hop. In this paper, we present the Holistic Multi-modal Memory Network (HMMN) framework which fully considers the interactions between different input sources (multi-modal context, question) in each hop. In addition, it takes answer choices into consideration during the context retrieval stage. Therefore, the proposed framework effectively integrates multi-modal context, question, and answer information, which leads to more informative context retrieved for question answering. Our HMMN framework achieves state-of-the-art accuracy on MovieQA dataset. Extensive ablation studies show the importance of holistic reasoning and contributions of different attention strategies.
\end{abstract}


%
\IEEEpeerreviewmaketitle

\section{Introduction}

With recent tremendous progress in computer vision and natural language processing, increasing attention has been drawn to the joint understanding of the visual and textual semantics. Related topics such as image-text retrieval~\cite{ma2015multimodal,huang2017instance,wang2018learning}, image/video captioning~\cite{you2016image,pan2016hierarchical,rennie2017self,nian2017learning,zhou2018end,zhang2019more}, visual question answering (VQA)~\cite{antol2015vqa,goyal2017making,johnson2017clevr,jain2018two} have been intensely explored. In particular, VQA remains a relatively challenging task, and it requires understanding of a given image to answer the question.

Due to the multi-modal nature of the world, developing question answering (QA) systems that are able to attain an understanding of the world based on multiple sources of information is a natural next step. Several QA datasets incorporating multiple data modalities have recently been developed  towards this end~\cite{kembhavi2017you,MovieQA,kim2017deepstory}. In this work, we focus on Movie question answering (MovieQA)~\cite{MovieQA}, which requires systems to demonstrate story comprehension by successfully answering multiple choice questions relating to videos and subtitles taken from movies.

A key challenge in multi-modal QA is to integrate information from different data sources. In the context of MovieQA, both query-to-context attention and inter-modal attention between videos and subtitles should be considered. Recently developed methods have adopted the classic strategies of early-fusion~\cite{na2017read} and late-fusion~\cite{MovieQA}, both of which have their limitations. Early-fusion of different modalities may limit the ability to pick up meaningful semantic correlations due to the increased noise at the feature level, while late-fusion does not allow for cross-referencing between modalities to define the higher level semantic features.
Wang \emph{et al.}~\cite{Wang2018} proposed to utilize inter-modal attention for MovieQA. However, their method does not fully integrate the input data, where different attention stage considers a different subset of interactions between the question, videos, subtitles for context retrieval.

Moreover, answer choices are only considered at the final step of the system where they are matched against an integrated representation of the input data, hence the useful contexts among answer choices are not effectively utilized to determine the relevant parts of the input data.

\begin{figure*}[t]
\centering
\includegraphics[width=1\textwidth]{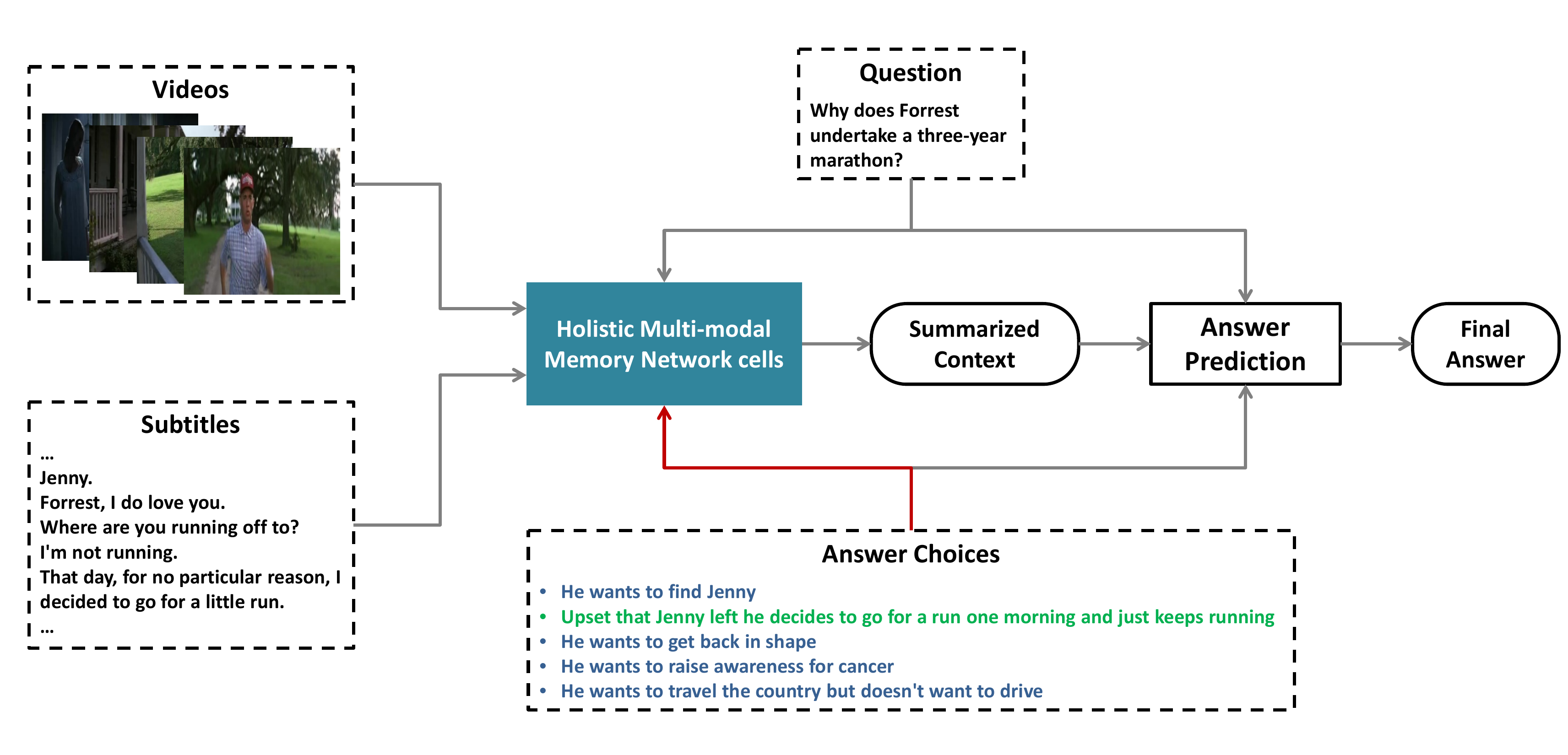}
\caption{Illustration of our proposed multi-modal feature learning framework. Our Holistic Multi-modal Memory Network cells holistically fuse multi-modal context (videos, subtitles), the question, as well as the answer choices. This framework jointly takes inter-modal and query-to-context attentions into consideration in each attention hop, and incorporates answer choices in both context retrieval and answer prediction stages.}
\label{fig:framework}
\end{figure*}

To address these limitations, we propose the Holistic Multi-modal Memory Network (HMMN) framework.
Firstly, our framework employs both inter-modal and query-to-context attention mechanisms for effective data integration in each hop. Specifically, our attention mechanism holistically investigates videos, subtitles, question, answer choices to obtain a summarized context in each attention hop, which is different from existing methods that only consider a subset of interactions in each hop. Hence, query-to-context relationship is jointly considered while modeling the multi-modal relationship between context. Secondly, our framework considers answer choices during not only answer prediction stage but also context retrieval stage. Utilizing answer choices to hone in on relevant information is a common heuristic used by students when taking multiple-choice tests. Analogously, we thought this would help on the QA task by restricting the set of inputs considered by the model thus helping it sieve out signal from the noise. Specifically, the retrieved answer-aware context should match the answer choice for correct answers. Otherwise, the resultant context may convey different semantic meaning from the answer choice.

Given this holistically modeling, our HMMN framework achieves state-of-the-art performance on both validation and test sets for video-based movie question answering. Ablation studies confirm the utility of incorporating answer choices during the context retrieval stage, which show that incorporating the answer information contributes to the context retrieval process. In addition, we include analyses of different attention mechanisms (query-to-context attention, inter-modal attention, intra-modal self attention) and confirm the importance of holistic reasoning.

The rest of this paper is organized as follows. Section II introduces related works on multi-modal question answering and visual question answering as well. Section III describes the HMMN method in detail. Section IV presents the experimental results together with analyses of different attention mechanisms. Section V concludes the paper.

\section{Related Work}
In this section, methods of visual question answering and multi-modal question answering which are closely related to our method are introduced.

\subsection{Visual Question Answering}

Besides the significant progress from computer vision and natural language processing, the emergence of large visual question answering (VQA) datasets~\cite{vqa_v1,balanced_vqa_v2,johnson2017clevr} also leads to the popularity of VQA task. Early works about VQA task use the holistic full-image feature to represent the visual context. For example, Malinowski \emph{et al.}~\cite{malinowski2015ask} proposed to feed the Convolutional Neural Network (CNN) image features and the question features together into a long short-term memory (LSTM) network and train an end-to-end network. Later, quite a few works have used attention mechanism to pay attention to certain parts of the image, where the alignment between image patches~\cite{xu2016ask,zhu2016visual7w} or region proposals~\cite{shih2016look} with the words in the question have been explored. Several attention mechanisms of connecting the visual context and the question have been proposed~\cite{xiong2016dynamic,lu2016hierarchical}. For example, Lu \emph{et al.}~\cite{lu2016hierarchical} presented a co-attention framework which considers visual attention and question attention jointly. Wu \emph{et al.}~\cite{wu2018image} proposed to incorporate high-level concepts and external knowledge for image captioning and visual question answering. Answer attention was investigated for the grounded question answering task~\cite{hu2017answer}. Grounded question answering is a special type of VQA, which is to retrieve an image bounding box from the candidate pool to answer the textual question. This method models the interaction between the answer candidates and the question and learns the answer-aware summarization of the question, while our method models the interaction between the answer choices and the context to retrieve more informative context.

\subsection{Multi-modal Question Answering}
In contrast to VQA, which only involves visual context, multi-modal question answering takes multiple modalities as context, and has attracted great interest. Kembhavi \emph{et al.}~\cite{kembhavi2017you} presented the Textbook Question Answering (TextbookQA) dataset that consists of lessons from middle school science curricula with both textual and diagrammatic context. In~\cite{kim2017deepstory}, PororoQA dataset was introduced, which is constructed from children cartoon Pororo with video, dialogue, and description. Tapaswi \emph{et al.}~\cite{MovieQA} introduced the movie question answering (MovieQA) dataset which aims to evaluate the story understanding from both video and subtitle modalities. In this paper, we focus on the MovieQA dataset, and related approaches are discussed as follows.

Most methods proposed for the MovieQA dataset are based on the End-to-end Memory Network~\cite{sukhbaatar2015end}, which was originally proposed for the textual question answering task. In~\cite{MovieQA}, they proposed a straightforward extension of the End-to-end Memory Network~\cite{sukhbaatar2015end} to multi-modal scenario on MovieQA. In particular, answer is predicted based on each modality separately, and late fusion is performed to combine the answer prediction scores from two modalities. Na \emph{et al.}~\cite{na2017read} proposed another framework based on the End-to-end Memory Network~\cite{sukhbaatar2015end}. Their framework has read and write networks that are implemented by convolutional layers to model sequential memory slots as chunks. Context from textual and visual modalities are early fused as the input to the write network. Specifically, compact bilinear pooling~\cite{gao2016compact} is utilized to obtain the joint embeddings with subshot and sentence features. Deep embedded memory networks (DEMN) model was introduced by~\cite{kim2017deepstory}, where they aim to reconstruct stories from a joint stream of scene and dialogue. However, their framework is not end-to-end trainable, as their method makes a hard context selection. Liang \emph{et al.}~\cite{liang2018focal} presented a focal visual-text attention network which captures correlation between visual and textual sequences for the personal photos and descriptions in the MemexQA dataset~\cite{jiang2017memexqa} and applied this method to the MovieQA dataset. Wang \emph{et al.}~\cite{Wang2018} proposed a layered memory network with two-level correspondences. Specifically, the static word memory module corresponds words with regions inside frames, and the dynamic subtitle memory module corresponds sentences with frames.
However, visual modality is used to attend to the textual modality which is dynamically updated by different strategies. Interactions between the question, videos, subtitles are not holistically considered in each attention stage.

\section{Methodology}
We will first introduce the notations. Then, we will introduce the End-to-end Memory Network (E2EMN)~\cite{sukhbaatar2015end}. After that, Holistic Multi-modal Memory Network (HMMN) cell will be introduced together with the prediction framework.

\subsection{Notations}
We assume the same feature dimension for subtitle sentences and frames. In MovieQA dataset, each question is aligned with several relevant video clips. We obtain features for frames and sentences following~\cite{Wang2018}.

Let $S \in {\mathbb{R}^{d \times m}}$ denote the subtitle modality, where $d$ is the dimension of the feature vectors and $m$ is number of subtitle sentences. For the subtitle modality, we not only gather the subtitle sentences within the relevant video clips, but also incorporate nearby (in time) subtitle sentences to make use of the contextual information. The dimension of the word2vec features for words in subtitles is $d_w$. The word2vec representation of each word is projected to $d$-dim with a projection matrix $W_1 \in {\mathbb{R}^{d_w \times d}}$. Then, a mean-pooling is performed among all words in each sentence to get the sentence representation.

Similarly, $V \in {\mathbb{R}^{d \times n}}$ represents the video modality, where $d$ is the dimension of the feature vectors and $n$ is number of frames. We select a fixed number of frames from the relevant video clips for each question. Frame-level representations are generated by investigating attention between regional features and word representations in the vocabulary, where $W_2 \in {\mathbb{R}^{d_r \times d_w}}$ is utilized to project the regional VGG~\cite{simonyan2015very} features to $d_w$-dim to match the dimension of word representations. Here $d_r$ is the dimension of the regional features. With regional features represented by the vocabulary word features, frame-level representations are generated with average pooling followed by a projection with $W_1$. We refer the reader to the original paper~\cite{Wang2018} or their released code for more details.

\begin{figure*}[t]
\centering
\includegraphics[width=1\textwidth]{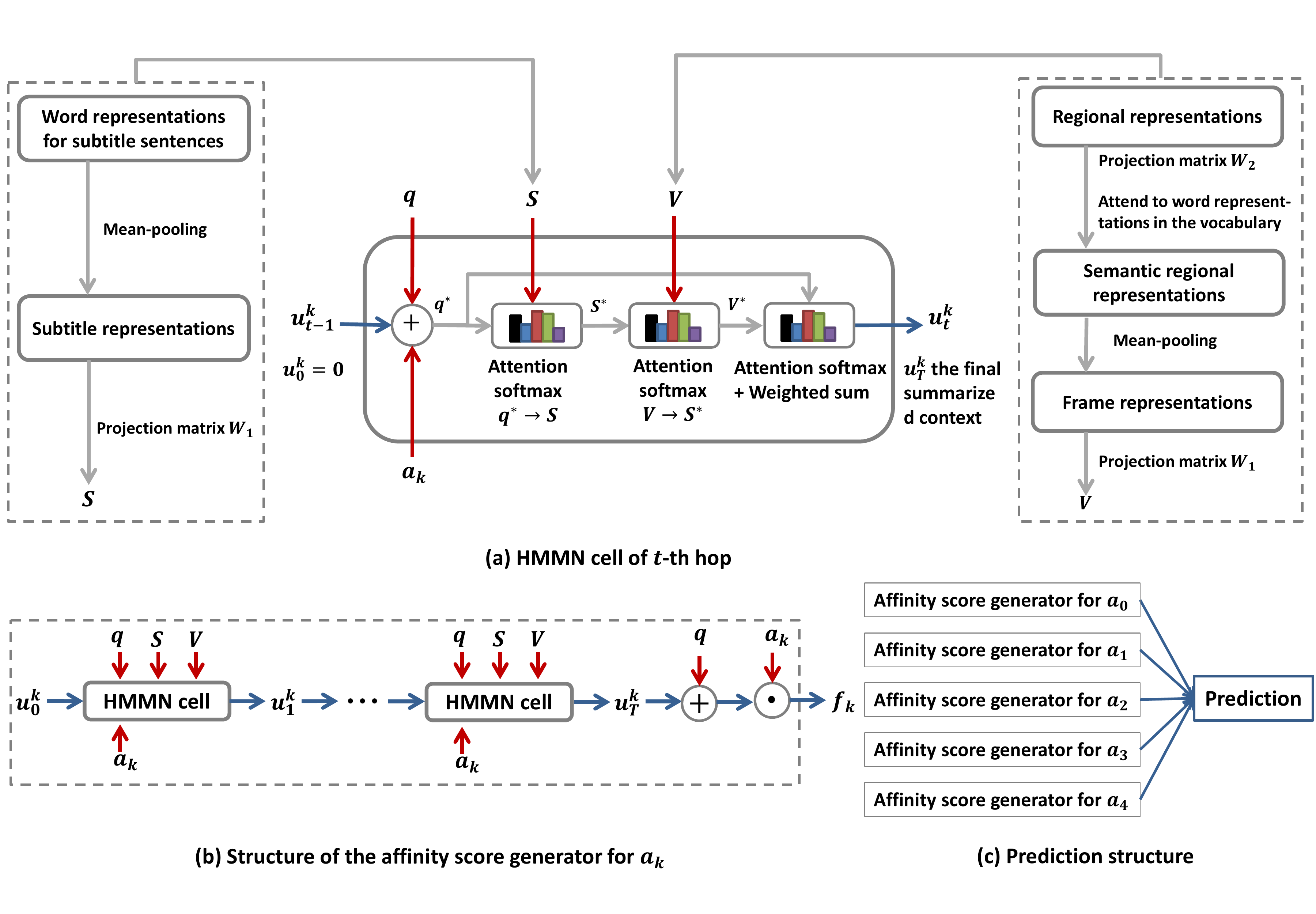}
\caption{Illustration of (a) the HMMN cell of $t$-th hop, (b) affinity score generator for answer choice $a_k$, and (c) the prediction structure. In (b), $T$ is the number of hops, which also denotes the number of stacked HMMN cells. $\oplus$ denotes the element-wise addition and $\odot$ denotes the element-wise multiplication.}
\label{fig:HMMN}
\end{figure*}

The question and answer choices are represented in the same way as subtitle sentences. The question is represented as a vector $q \in {\mathbb{R}^{d}}$. Answer choices for each question are represented as $A= \left[a_0, a_1, a_2, a_3,a_4\right] \in {\mathbb{R}^{d \times 5}} $, where each answer choice is encoded as $a_k\in {\mathbb{R}^{d}}$. In the whole framework, only $W_1$ and $W_2$ are learnable. The structures to generate representations for the subtitle and video modalities are shown in Fig.~\ref{fig:HMMN}(a).

\subsection{End-to-end Memory Network}
The End-to-end Memory Network (E2EMN)~\cite{sukhbaatar2015end} is originally proposed for a question answering task where the aim is to pick the most likely word from the vocabulary as answer according the textual context. In~\cite{MovieQA}, E2EMN is adapted to multi-modal question answering with multi-choice answers. In particular, scores from two modalities are late-fused to make the final prediction. As E2EMN is designed for textual question answering, this method only deals with context from a single modality. Here we use the subtitle modality $S$ for explanation.

In E2EMN, input features of context $S$ are treated as memory slots. With both memory slots and query (question is used as query here) as input, a summary of context is derived according to the relevance between the query and memory slots.
In particular, the match between query $q$ and each memory slot is calculated with the inner product followed by a softmax:
\begin{equation}\label{eq:q1}
{\alpha _i} = softmax ({{ q}^T}{S_{:i}})
\end{equation}
where ${\alpha _i}$ indicates the importance of $i$-th subtitle sentence to the query. The summarized context $u$ is computed as the weighted sum of subtitle sentence features based on ${\alpha _i}$:
\begin{equation}\label{eq:q2}
u = \sum\limits_{i = 1}^m {{\alpha _{i}}{S_{:i}}}
\end{equation}

Then, the answer prediction is made by comparing the answer choice $a_i$ with the sum of query representation $q$ and the summarized context $u$:
\begin{equation}\label{eq:q3}
p = softmax({(q + u)^T}A)
\end{equation}
where $p \in {\mathbb{R}^{5}}$ is the confidence vector. Here $5$ is the number of answer choices for MovieQA. The process to derive the summarized context can be performed in multiple hops, where the output of one layer can be used as part of the query of the next layer.

\subsection{Holistic Multi-modal Memory Network (HMMN)}
Different from  E2EMN, our HMMN framework takes multi-modal context as input. HMMN framework investigates interactions between multi-modal context and the question jointly. By doing this, query-to-context relationship is jointly considered while modeling the multi-modal relationship between context. In addition, it not only exploits answer choices for answer prediction but also in the process of summarizing the context from multiple modalities.

The inference process is performed by stacking small building blocks, called HMMN cell. The structure of HMMN cell is shown in Fig.~\ref{fig:HMMN}(a). The process to generate $S$ and $V$ is only for illustration, while our main contribution lies in the HMMN structure. Each HMMN cell takes as input the question, one answer choice, context from videos and subtitles, and derives the answer-aware summarized context. We call this process as one hop of reasoning. Let $u_t^k$ be the output of the $t$-th reasoning hop with respect to answer choice $k$. The output of the $t$-th hop will be utilized as the input of the $(t+1)$-th hop.
\subsubsection{Involving answers in context retrieval}

The HMMN cell incorporates the answer choice as a part of the query in the context retrieval stage. The query involving $k$-th answer choice for the $t$-th hop is calculated by combining the output of previous hop $u_{t-1}^k$, the question $q$, answer choice $a_k$:
\begin{equation}
q^* = u_{t-1}^k+a_k+\lambda q
\end{equation}
where $\lambda$ is a tradeoff parameter between the question and the rest of the query.

The intuition of incorporating answer choices in the context retrieval stage is to mimic behaviors of students who take a reading test with multi-choice questions. When the context is long and complicated, a quick and effective way to answer the question is to locate relevant information with respect to each answer choice. For one answer choice, if the retrieved answer-aware context conveys similar ideas with the answer choice, it tends to be the correct answer. Alternatively, if the retrieved context has a different semantic meaning, the answer is likely to be wrong.

\subsubsection{Holistically considering different attention mechanisms in each hop}
Instead of only taking a subset of interactions between the query and multi-modal context, our framework jointly considers inter-modal and query-to-context attention strategies in each hop.

The HMMN cell takes the query $q^*$ to gather descriptive information from multi-modal context, where interactions between the question, answer choices, videos, subtitles are exploited holistically.
In particular, we utilize the updated query to highlight the relevant subtitle sentences in $S$ by performing the query-to-context attention (denoted as $(q^* \to S)$). The resulted re-weighted subtitle modality is represented as $S^*$:

\begin{equation}
\begin{aligned}
& {\delta _i} = softmax ({{ q^*}^T}{S_{:i}})\\
& ~~~~~~{{S}_{:i}^*} = { \delta_i}{S_{:i}}\\
\end{aligned}
\end{equation}
where more relevant subtitle sentences are associated with large weights.

Inter-modal attention reasoning is applied by using the video modality $V$ to attend to the subtitle modality $S$ (denoted as ($V \to S^*$)), which aims to generate the subtitle-aware representations for frames as $V^*$. Each frame is represented with the weighted sum of all the subtitle sentence features according to the relevance:
\begin{equation}
\begin{aligned}
& ~{\epsilon _{ij}} = {V_{:i}}^T{S_{:j}^*} \\
& {{V}_{:i}^*} = \sum\limits_{j = 1}^m {{\epsilon _{ij}}{S_{:j}^*}} \\
\end{aligned}
\end{equation}
The resulted $V^*$ can be summarized with respect to the query $q^*$ as the hop output. In particular, the $t$-hop summarized context with respect to the $k$-th answer choice is caculated as $u^k_t$:
\begin{equation}
\begin{aligned}
& {\zeta _i} = softmax ({{ q^*}^T}{V_{:i}^*}) \\
& ~~~~~u^k_t = \sum\limits_{i = 1}^n {{\zeta _{i}}{V_{:i}^*}} \\
\end{aligned}
\end{equation}

In each reasoning hop, the output of previous hop, the answer choice, the question, multi-modal context are holistically integrated. The reason of using $V$ to attend to $S$ (not using $S$ to attend to $V$) is that, the subtitle modality is more informative than the video modality for the MovieQA task. Typically, the subtitle modality includes descriptions of the story such as character relationships,
story development. By attending to $S$, the feature representations in $S$ will be used to form the summarized context. The re-weighted $S$ acts as an information screening step to derive more informative representations for the subtitle modality.

\subsubsection{Predicting answer with affinity scores}

It is shown in the original E2EMN that multiple hops setting yields improved results. We stack the HMMN cells to do $T$ hops of reasoning. Given the final summarized context $u^k_T$ with respect to answer choice $a_k$, an affinity score $f_k$ for $a_k$ is generated. This score is derived by comparing the sum of the question and answer-aware summarized context with the answer choice as:
\begin{equation}
f_k = {(q + u_T^k)^T}a_k
\end{equation}
This score indicates whether the retrieved context has the consistent semantic meaning with the answer choice. The structure of generating the affinity score is shown in Fig.~\ref{fig:HMMN}(b). Then the affinity scores for all the answer choices will be passed to a softmax function to get the final answer prediction as shown in Fig.~\ref{fig:HMMN}(c), and the cross-entropy loss is minimized with the standard stochastic gradient descent. This indicates that if one answer choice matches with the answer-aware summarized context, it is likely to be the correct answer.

\section{Experiments}

\subsection{Dataset and Setting}
The MovieQA dataset~\cite{MovieQA} consists of 408 movies and 14,944 questions. Diverse sources of information are collected including video clips, plots, subtitles, scripts, and Descriptive Video Service (DVS). Plot synopses from Wikipedia are utilized to generate questions. For multi-modal question answering task with videos and subtitles, there are 6,462 questions with both videos clips and subtitles from 140 movies. We follow the public available train, validation, test split. The accuracy of multi-choice questions is measured.

\begin{table}
\caption{Performances of HMMN structures w/ and w/o answer attention with different numbers of layers on the validation set.} \label{tab:layer}
\begin{center}
\begin{tabular}{p{5.5cm}|>{\centering\arraybackslash}p{2cm}}
\hline
~~~~~~~~~~~~~~~~~~~~~~~~~Method & Accuracy (\%) \\
\hline\hline
HMMN (1 layer)  ~w/o answer attention& 43.35\\
\hline
HMMN (2 layers) w/o answer attention& \textbf{44.47} \\
\hline
HMMN (3 layers)  w/o answer attention& 44.24\\
\hline\hline
HMMN (1 layer) & 45.71\\
\hline
HMMN (2 layers) & \textbf{46.28} \\
\hline
HMMN (3 layers)  & 44.13\\
\hline
\end{tabular}
\end{center}
\end{table}

\begin{figure}
\centering
\includegraphics[width=0.48\textwidth]{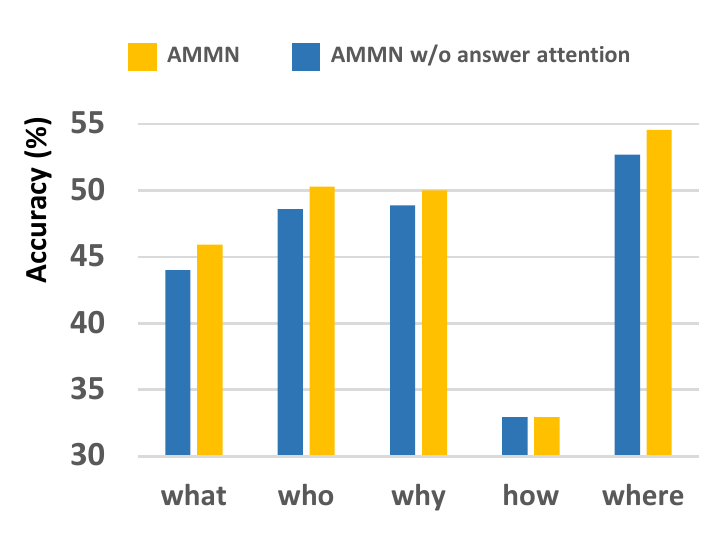}
\caption{Comparison between HMMN with and without answer attention for different question types.}
\label{fig:HMMN_wo}
\end{figure}

\subsection{Implementation Details}
For the subtitle modality, we consider the subtitle sentences which fall into the time interval derived by extending the starting and ending time points of video clips by 300 seconds. For the video modality, 32 frames are selected from the relevant video clips following~\cite{Wang2018}. We use the word2vec representations provided by~\cite{MovieQA}. The dimension of the word2vec representation $d_w$ is 300.  The dimension of the regional features from `pool5' of VGG-16 $d_r$ is 512. 10\% of the training examples are kept as the development set. The batch size is 8. The learning rate is set to 0.005. The tradeoff parameter $\lambda$ is set to be 0.45. The dimension of features $d$ is set to 300. Our model is trained up to 50 epochs, and early stopping is performed.

\subsection{Quantitative Analysis}

Table~\ref{tab:layer} presents results for HMMN structures with and without answer attention with different numbers of layers, where ``with answer attention" means considering answer choices when retrieving the context. We can see that by incorporating the answer choices in the context retrieval stage, the performance is significantly improved.
And 2-layer structures achieve the best performance, thus the number of layers which is also the number of hops $T$ is set to 2. Fig.~\ref{fig:HMMN_wo} shows the the comparison of HMMN w/ and w/o answer attention for different question types, with the starting word as `what', `who', `why', `who', `where'. It can be seen that HMMN framework performs consistently better than the HMMN framework w/o answer attention.

Table~\ref{tab:comparison} shows the comparison with state-of-the-art methods. We compare with~\cite{MovieQA,na2017read,kim2017deepstory,liang2018focal,Wang2018}. It can be seen that our proposed method significantly outperforms all four state-of-the-art methods on both validation and test sets.

In particular,~\cite{MovieQA} performs a late fusion.~\cite{na2017read} conducts an early fusion. Neither late fusion nor early fusion can well exploit the relationship between modalities, which results in suboptimal results.~\cite{kim2017deepstory} is not end-to-end trainable.~\cite{liang2018focal} assumes sequences of the visual and textual representations have the same length, which is not true for MovieQA. Similar with~\cite{na2017read},~\cite{liang2018focal} cuts off visual information of those frames without accompanying subtitles.~\cite{Wang2018} takes the multi-modal relationship into consideration, however, in each attention stage, a different subset of interactions between the question, videos, subtitles are considered. In comparison, our HMMN framework w/o answer attention holistically incorporates the output of previous hop, the question, videos, subtitles in each hop, which leads to superior performance.

\begin{table}[t]
\caption{Comparison of state-of-the-art methods on validation and test sets.} \label{tab:comparison}
\begin{center}
\begin{tabular}{p{3.6cm}|>{\centering\arraybackslash}p{2cm}|>{\centering\arraybackslash}p{2cm}}
\hline
~~~~~~~~~Method & Accuracy on Val (\%) & Accuracy on Test (\%)\\
\hline\hline
Tapaswi \emph{et al.}~\cite{MovieQA} & 34.20 & -\\
\hline
Na \emph{et al.}~\cite{na2017read} & 38.67 & 36.25 \\
\hline
Kim \emph{et al.}~\cite{kim2017deepstory} & 44.7 & 34.74 \\
\hline
Liang \emph{et al.}~\cite{liang2018focal} & 41.0 & 37.3\\
\hline
Wang \emph{et al.}~\cite{Wang2018} & 42.5 & 39.03\\
\hline\hline
Our HMMN framework & &\\~ w/o answer attention & 44.47&41.65\\
\hline
Our HMMN framework & \textbf{46.28}& \textbf{43.08}\\
\hline
\end{tabular}
\end{center}
\end{table}

\begin{table}
\caption{Performance of baselines with different attention strategies on the validation set.} \label{tab:baseline}
\begin{center}
\begin{tabular}{p{5.5cm}|>{\centering\arraybackslash}p{2cm}}
\hline
Method & Accuracy (\%) \\
\hline\hline
$V$ & 37.69\\
\hline
$S$ & 39.62 \\
\hline
$V'$~~$(q \to V)$ Query-to-context Attention  &  37.92\\
\hline
$S'$~~$(q \to S)$ Query-to-context Attention  &  40.86\\
\hline
$\bar V$~~$(V \to S)$ Inter-modal Attention&  42.73\\
\hline
$\bar S$~~$(S \to V)$ Inter-modal Attention&  35.10\\
\hline
$\hat V$~~$(V \to V)$ Intra-modal Self Attention&  37.92\\
\hline
$\hat S$~~$(S \to S)$ Intra-modal Self Attention &  40.29\\
\hline
\end{tabular}
\end{center}
\end{table}

\subsubsection{Ablation study}

In this paper, we also explore different attention mechanisms. Effects of different attention strategies are investigated.

\setlength{\parindent}{0cm}
\textbf{(i) Query-to-context Attention ~~~($q \to S$) ~~~$S'$}
\setlength{\parindent}{0.35cm}

Query-to-context attention indicates which memory slots are more relevant to the query. Here $S$ of subtitle modality is used as the context for illustration. With the calculated similarity between the query and each memory slot in Eq.~\ref{eq:q1}, more relevant subtitle sentences can be highlighted with:
\begin{equation}
{{S'}_{:i}} = {\alpha _i}{S_{:i}}
\end{equation}

\begin{table}
\caption{Performance of baselines exploring higher-level Inter-modal Attention on the validation set.} \label{tab:baseline_2}
\begin{center}
\begin{tabular}{>{\centering\arraybackslash}p{1.2cm}|>{\centering\arraybackslash}p{2cm}||>{\centering\arraybackslash}p{1.2cm}|>{\centering\arraybackslash}p{2cm}}
\hline
Method & Accuracy (\%)& Method & Accuracy (\%) \\
\hline\hline
$V \to S$ & 42.73 & $S \to V$ & 35.10\\
\hline
$V \to S'$ & \textbf{43.35} & $S' \to V$ & 35.21\\
\hline
$V \to \bar S$ & 37.47 & $\bar S \to V$ & 35.10\\
\hline
$V \to \hat S$ & 41.08 & $\hat S \to V$ & 35.10\\
\hline\hline

$V' \to S$ & 43.12 & $S \to V'$ & 35.21\\
\hline
$V' \to S'$ & \textbf{43.35}& $S' \to V'$ & 35.44\\
\hline
$V' \to \bar S$ & 38.14 & $\bar S \to V'$ & 35.32\\
\hline
$V' \to \hat S$ & 37.02 & $\hat S \to V'$ & 35.44\\
\hline\hline

$\bar V \to S$ & 41.76 & $S \to \bar V$ & 40.29\\
\hline
$\bar V \to S'$ & 41.08 & $S' \to \bar  V$ & 40.18\\
\hline
$\bar V \to \bar S$ & 39.84 & $\bar S \to \bar V$ & 40.85\\
\hline
$\bar V \to \hat S$ & 37.47 & $\hat S \to \bar V$ & 38.60\\
\hline\hline

$\hat V \to S$ & 43.12 & $S \to \hat V$ & 34.55\\
\hline
$\hat V \to S'$ & \textbf{43.35} & $S' \to \hat V$ & 35.77\\
\hline
$\hat V \to \bar S$ & 37.58 & $\bar S \to \hat V$ & 35.10\\
\hline
$\hat V \to \hat S$ & 38.15 & $\hat S \to \hat V$ & 34.98\\
\hline
\end{tabular}
\end{center}
\end{table}

This process is denoted as $q \to S$, and the re-weighted memory slots are represented as $S'$.

\begin{figure*}[t]
\centering
\includegraphics[width=1\textwidth]{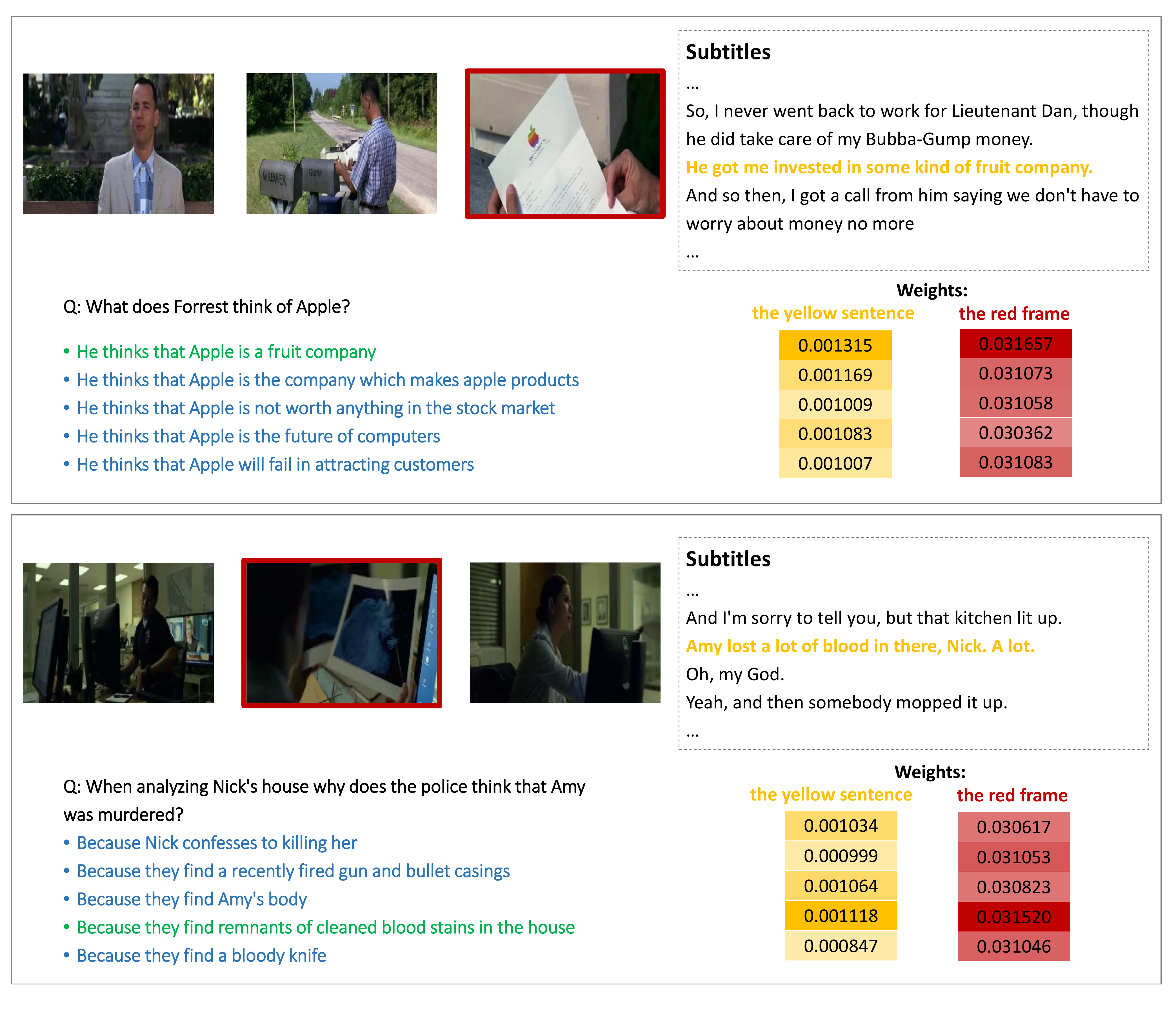}
\caption{Visualization of weights of two success cases in the HMMN framework. The correct answer choice is colored in green. Weights of relevant frame and subtitle sentence (highlighted in red and yellow) are shown. When using the correct answer choice to search for the relevant context, the relevant frame and sentence are associated with higher weights than those for other answer choices. The correct answer choice is in green. }
\label{fig:vis}
\end{figure*}

\begin{figure*}[t]
\centering
\includegraphics[width=0.92\textwidth]{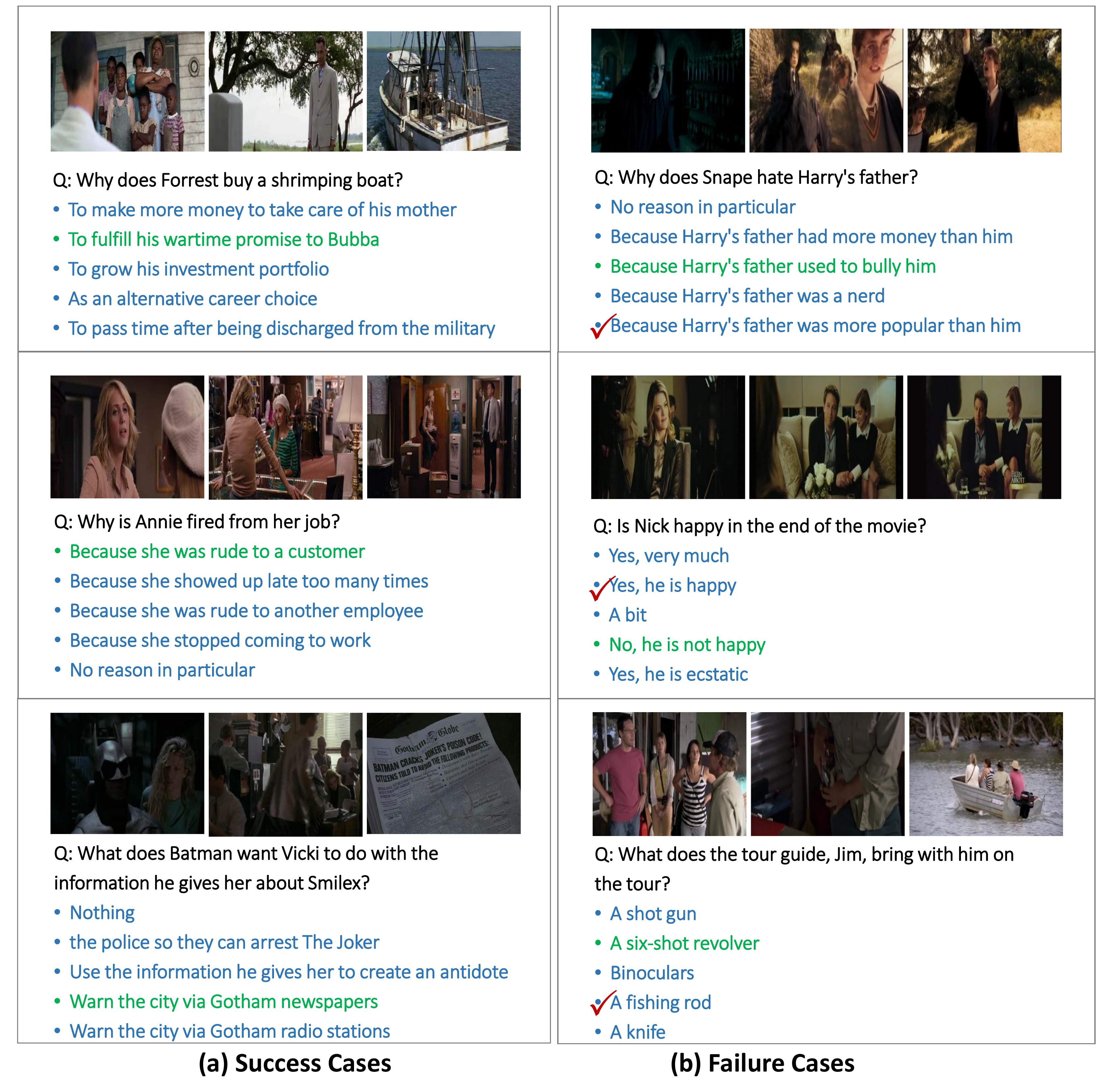}
\caption{Success and failure cases. The correct answer choice is in green and the mistake made by our framework is marked in red. Failure cases indicate the necessity of exploiting common sense knowledge and object detection results.}
\label{fig:suc_fail}
\end{figure*}

\setlength{\parindent}{0cm}
\textbf{(ii) Inter-modal Attention} ~~~($S \to V$)~~~$\bar S$
\setlength{\parindent}{0.35cm}

Inter-modal attention from $S$ to $V$ indicates that, for each subtitle sentence, we intent to find the most relevant frames. The retrieved frame features will be fused to represent the subtitle sentence. The output can be interpreted as the video-aware sentence representation.

First, the coattention matrix between frames and subtitle sentences can be defined as:
\begin{equation}\label{eq:q5}
{\beta _{ij}} = {S_{:i}}^T{V_{:j}}
\end{equation}
where ${\beta _{ij}}$ indicates the relevance between $j$-th frame and $i$-th subtitle sentence. Then $i$-th subtitle sentence can be represented by the weighted sum of all frames based on ${\beta _{ij}}$:
\begin{equation}\label{eq:q6}
{{\bar S}_{:i}} = \sum\limits_{j = 1}^n {{\beta _{ij}}{V_{:j}}}
\end{equation}
The resulted representation for subtitle modality $\bar S$ is of the same size as $S$. Similarly, the result for ($V \to S$) is $\bar V$.

\setlength{\parindent}{0cm}
\textbf{(iii) Intra-modal Self Attention} ~~~($S \to S$)~~~$\hat S$
\setlength{\parindent}{0.35cm}

Self attention has shown its power in tasks such as question answering and machine translation~\cite{hudson2018compositional,vaswani2017attention}. The intuition is that contextual information among other memory slots can be exploited. Similar with the inter-modal attention, the coattention between different memory slots in the same modality is calculated as:
\begin{equation}
{\gamma _{ij}} = I(i \ne j){S_{:i}}^T{S_{:j}}
\end{equation}
Noted that the correlation between one sentence with itself is set to zero. The resulted representation with self-attention is ${\hat S}$. Each subtitle sentence will be represented by the weighted sum of features of all the subtitle sentences based on $\gamma_{ij}$:
\begin{equation}
{{\hat S}_{:i}} = \sum\limits_{j = 1}^m {{\gamma _{ij}}{S_{:j}}}
\end{equation}

With derived $S$, $S'$, $\bar S$, $\hat S$, $V$, $V'$, $\bar V$, $\hat V$, each of them can be treated as memory slots in the original E2EMN. Similarly, Eq.~\ref{eq:q1}, Eq.~\ref{eq:q2} and Eq.~\ref{eq:q3} can be applied to predict the answer.

Table~\ref{tab:baseline} shows results of baselines with different attention strategies. For $S$, $S'$, $\bar S$, $\hat S$, $V$, $V'$, $\bar V$, $\hat V$, each baseline treats one of them as memory slots in the original E2EMN to predict the answer. The baseline $S$ performs better than $V$, as the subtitle modality contains more informative descriptions of character relationships and story development. By using each attention strategy to obtain the enhanced representations, the performance improvement is achieved. Particularly, the inter-modal attention $(V \to S)$ brings a significant improvement.

To explore higher-level inter-modal attention, we use $V$, $V'$, $\bar V$, $\hat V$ to attend to $S$, $S'$, $\bar S$, $\hat S$, and vise versa with Eq.~\ref{eq:q5} and Eq.~\ref{eq:q6}. The results are shown in Table~\ref{tab:baseline_2}. Typically, the baselines on the left side perform better than ones on the right side. As mentioned in the methodology section, it is because when we use $V \to S$ attention, the video modality will be represented by the subtitle features, which are more descriptive. According to this observation, when designing the structure of attention mechanism for any general multi-modal tasks, using the less discriminative modalities as clues to attend to the more discriminative modalities tends to achieve better performance. We can observe that the intra-modal self attention does not bring much improvement to this task. Baselines of using the video modality to attend to re-weighted subtitle modality ($V \to S'$) perform considerably well, and our 1-layer HMMN framework w/o answer attention degenerates to the $V \to S'$ baseline.

\subsection{Qualitative Analysis}
To demonstrate that relevant context can be well captured by the HMMN framework, we visualize the attention weights of frames and subtitles of two success cases in Fig.~\ref{fig:vis}(a). The observed relevant frame and subtitle are highlighted in red and yellow respectively. For the question ``What does Forrest think of Apple?", the first answer choice is correct. Following attention weights of the second layer are visualized: 1) the attention weight of the subtitle sentence with respect to the answer choice and question; 2) the attention weight of the frame with respect to the answer choice and question. By using the first answer choice to retrieve the context, the relevant frame and subtitle sentence are associated with larger weights compared to those of other choices. Thus the key information can be captured and a high affinity score will be generated. On the other hand, the HMMN framework w/o answer attention picks the second answer choice as the correct one. This is because that the word ``Apple" is not mentioned in the subtitle, thus by using the question only to retrieve the information, important context about ``Apple" is missed. Similar remarks can be made for the second example question in Fig.~\ref{fig:vis}(a), the relevant frame and subtitle sentence are given high weights.

Fig.~\ref{fig:vis}(b) shows 3 typical failure cases. In the first example, although both video and subtitles contain the information that Harry's father made fun of Snape, it is difficult to associate them with the word ``bully" that has a high-level semantic meaning, where common sense knowledge reasoning is required. The second example is from ``Gone girl". Although the husband is not happy with the main character Amy, the actors act as having a happy ending. This question also requires common sense to answer. In the third example, the revolver appears in the frames, but is not mentioned in the subtitles. Although we use conventional CNN features to generate frame-level representations, the associations between visual patterns and object labels are not enforced during training. More success and failure cases can be found in Fig.~\ref{fig:suc_fail}.

\textbf{Future work}: Failure cases indicate the necessity of exploiting common sense knowledge and object detection results, which we leave for future work. High-level semantic information can be injected to the network by leveraging well-built knowledge graph, e.g. ConceptNet~\cite{liu2004conceptnet}. Object detector can capture more descriptive visual information than the current grid-based regions, and labels generated along with detected regions further enforce the correspondence of visual and textual information.

\section{Conclusion}
We presented a Holistic Multi-modal Memory Network framework that learns to answer questions with context from multi-modal data. In the proposed HMMN framework, we investigate both inter-modal and query-to-context attention mechanism to jointly model the interactions between multi-modal context and the question. In addition, we explore the answer attention by incorporating the answer choices in the context retrieval stage. Our HMMN framework achieves state-of-the-art results on the MovieQA dataset. We also presented a detailed ablation study for different attention mechanisms, which could provide guidance for future model design.

\ifCLASSOPTIONcaptionsoff
  \newpage
\fi



%

{
\bibliographystyle{IEEEtrans}
\bibliography{ref}

\begin{thebibliography}{10}
\providecommand{\url}[1]{#1}
\csname url@samestyle\endcsname
\providecommand{\newblock}{\relax}
\providecommand{\bibinfo}[2]{#2}
\providecommand{\BIBentrySTDinterwordspacing}{\spaceskip=0pt\relax}
\providecommand{\BIBentryALTinterwordstretchfactor}{4}
\providecommand{\BIBentryALTinterwordspacing}{\spaceskip=\fontdimen2\font plus
\BIBentryALTinterwordstretchfactor\fontdimen3\font minus
  \fontdimen4\font\relax}
\providecommand{\BIBforeignlanguage}[2]{{%
\expandafter\ifx\csname l@#1\endcsname\relax
\typeout{** WARNING: IEEEtranS.bst: No hyphenation pattern has been}%
\typeout{** loaded for the language `#1'. Using the pattern for}%
\typeout{** the default language instead.}%
\else
\language=\csname l@#1\endcsname
\fi
#2}}
\providecommand{\BIBdecl}{\relax}
\BIBdecl

\bibitem{antol2015vqa}
S.~Antol, A.~Agrawal, J.~Lu, M.~Mitchell, D.~Batra, C.~Lawrence~Zitnick, and
  D.~Parikh, ``Vqa: Visual question answering,'' in \emph{ICCV}, 2015.

\bibitem{vqa_v1}
S.~Antol, A.~Agrawal, J.~Lu, M.~Mitchell, D.~Batra, C.~L. Zitnick, and
  D.~Parikh, ``{VQA}: {V}isual {Q}uestion {A}nswering,'' in \emph{ICCV}, 2015.

\bibitem{gao2016compact}
Y.~Gao, O.~Beijbom, N.~Zhang, and T.~Darrell, ``Compact bilinear pooling,'' in
  \emph{CVPR}, 2016.

\bibitem{goyal2017making}
Y.~Goyal, T.~Khot, D.~Summers-Stay, D.~Batra, and D.~Parikh, ``Making the v in
  vqa matter: Elevating the role of image understanding in visual question
  answering,'' in \emph{CVPR}, 2017.

\bibitem{balanced_vqa_v2}
Y.~Goyal, T.~Khot, D.~Summers{-}Stay, D.~Batra, and D.~Parikh, ``Making the {V}
  in {VQA} matter: Elevating the role of image understanding in {V}isual
  {Q}uestion {A}nswering,'' in \emph{CVPR}, 2017.

\bibitem{hu2017answer}
J.~Hu, D.~Fan, S.~Yao, and J.~Oh, ``Answer-aware attention on grounded question
  answering in images,'' 2017.

\bibitem{huang2017instance}
Y.~Huang, W.~Wang, and L.~Wang, ``Instance-aware image and sentence matching
  with selective multimodal lstm,'' in \emph{CVPR}, 2017.

\bibitem{hudson2018compositional}
D.~A. Hudson and C.~D. Manning, ``Compositional attention networks for machine
  reasoning,'' \emph{arXiv preprint arXiv:1803.03067}, 2018.

\bibitem{jain2018two}
U.~Jain, S.~Lazebnik, and A.~Schwing, ``Two can play this game: Visual dialog
  with discriminative question generation and answering,'' in \emph{CVPR},
  2018.

\bibitem{jiang2017memexqa}
L.~Jiang, J.~Liang, L.~Cao, Y.~Kalantidis, S.~Farfade, and A.~Hauptmann,
  ``Memexqa: Visual memex question answering,'' \emph{arXiv preprint
  arXiv:1708.01336}, 2017.

\bibitem{johnson2017clevr}
J.~Johnson, B.~Hariharan, L.~van~der Maaten, L.~Fei-Fei, C.~L. Zitnick, and
  R.~Girshick, ``Clevr: A diagnostic dataset for compositional language and
  elementary visual reasoning,'' in \emph{CVPR}, 2017.

\bibitem{kembhavi2017you}
A.~Kembhavi, M.~Seo, D.~Schwenk, J.~Choi, A.~Farhadi, and H.~Hajishirzi, ``Are
  you smarter than a sixth grader? textbook question answering for multimodal
  machine comprehension,'' in \emph{CVPR}, 2017.

\bibitem{kim2017deepstory}
K.-M. Kim, M.-O. Heo, S.-H. Choi, and B.-T. Zhang, ``Deepstory: video story qa
  by deep embedded memory networks,'' in \emph{IJCAI}, 2017.

\bibitem{liang2018focal}
J.~Liang, L.~Jiang, L.~Cao, L.-J. Li, and A.~Hauptmann, ``Focal visual-text
  attention for visual question answering,'' in \emph{CVPR}, 2018.

\bibitem{liu2004conceptnet}
H.~Liu and P.~Singh, ``Conceptnet?a practical commonsense reasoning tool-kit,''
  \emph{BT technology journal}, vol.~22, no.~4, pp. 211--226, 2004.

\bibitem{lu2016hierarchical}
J.~Lu, J.~Yang, D.~Batra, and D.~Parikh, ``Hierarchical question-image
  co-attention for visual question answering,'' in \emph{NIPS}, 2016.

\bibitem{ma2015multimodal}
L.~Ma, Z.~Lu, L.~Shang, and H.~Li, ``Multimodal convolutional neural networks
  for matching image and sentence,'' in \emph{ICCV}, 2015.

\bibitem{malinowski2015ask}
M.~Malinowski, M.~Rohrbach, and M.~Fritz, ``Ask your neurons: A neural-based
  approach to answering questions about images,'' in \emph{ICCV}, 2015.

\bibitem{na2017read}
S.~Na, S.~Lee, J.~Kim, and G.~Kim, ``A read-write memory network for movie
  story understanding,'' in \emph{CVPR}, 2017.

\bibitem{nian2017learning}
F.~Nian, T.~Li, Y.~Wang, X.~Wu, B.~Ni, and C.~Xu, ``Learning explicit video
  attributes from mid-level representation for video captioning,''
  \emph{Computer Vision and Image Understanding}, vol. 163, pp. 126--138, 2017.

\bibitem{pan2016hierarchical}
P.~Pan, Z.~Xu, Y.~Yang, F.~Wu, and Y.~Zhuang, ``Hierarchical recurrent neural
  encoder for video representation with application to captioning,'' in
  \emph{CVPR}, 2016.

\bibitem{rennie2017self}
S.~J. Rennie, E.~Marcheret, Y.~Mroueh, J.~Ross, and V.~Goel, ``Self-critical
  sequence training for image captioning,'' in \emph{CVPR}, 2017.

\bibitem{shih2016look}
K.~J. Shih, S.~Singh, and D.~Hoiem, ``Where to look: Focus regions for visual
  question answering,'' in \emph{CVPR}, 2016.

\bibitem{simonyan2015very}
K.~Simonyan and A.~Zisserman, ``Very deep convolutional networks for
  large-scale image recognition,'' 2015.

\bibitem{sukhbaatar2015end}
S.~Sukhbaatar, J.~Weston, R.~Fergus \emph{et~al.}, ``End-to-end memory
  networks,'' in \emph{NIPS}, 2015.

\bibitem{MovieQA}
M.~Tapaswi, Y.~Zhu, R.~Stiefelhagen, A.~Torralba, R.~Urtasun, and S.~Fidler,
  ``{MovieQA: Understanding Stories in Movies through Question-Answering},'' in
  \emph{CVPR}, 2016.

\bibitem{vaswani2017attention}
A.~Vaswani, N.~Shazeer, N.~Parmar, J.~Uszkoreit, L.~Jones, A.~N. Gomez,
  {\L}.~Kaiser, and I.~Polosukhin, ``Attention is all you need,'' in
  \emph{NIPS}, 2017.

\bibitem{Wang2018}
B.~Wang, Y.~Xu, Y.~Han, and R.~Hong, ``Movie question answering: Remembering
  the textual cues for layered visual contents,'' in \emph{AAAI}, 2018.

\bibitem{wang2018learning}
L.~Wang, Y.~Li, J.~Huang, and S.~Lazebnik, ``Learning two-branch neural
  networks for image-text matching tasks,'' \emph{IEEE Transactions on Pattern
  Analysis and Machine Intelligence}, 2018.

\bibitem{wu2018image}
Q.~Wu, C.~Shen, P.~Wang, A.~Dick, and A.~van~den Hengel, ``Image captioning and
  visual question answering based on attributes and external knowledge,''
  \emph{IEEE transactions on pattern analysis and machine intelligence},
  vol.~40, no.~6, pp. 1367--1381, 2018.

\bibitem{xiong2016dynamic}
C.~Xiong, S.~Merity, and R.~Socher, ``Dynamic memory networks for visual and
  textual question answering,'' in \emph{ICML}, 2016.

\bibitem{xu2016ask}
H.~Xu and K.~Saenko, ``Ask, attend and answer: Exploring question-guided
  spatial attention for visual question answering,'' in \emph{ECCV}, 2016.

\bibitem{you2016image}
Q.~You, H.~Jin, Z.~Wang, C.~Fang, and J.~Luo, ``Image captioning with semantic
  attention,'' in \emph{CVPR}, 2016.

\bibitem{zhang2019more}
M.~Zhang, Y.~Yang, H.~Zhang, Y.~Ji, H.~T. Shen, and T.-S. Chua, ``More is
  better: Precise and detailed image captioning using online positive recall
  and missing concepts mining,'' \emph{IEEE Transactions on Image Processing},
  vol.~28, no.~1, pp. 32--44, 2019.

\bibitem{zhou2018end}
L.~Zhou, Y.~Zhou, J.~J. Corso, R.~Socher, and C.~Xiong, ``End-to-end dense
  video captioning with masked transformer,'' in \emph{CVPR}, 2018.

\bibitem{zhu2016visual7w}
Y.~Zhu, O.~Groth, M.~Bernstein, and L.~Fei-Fei, ``Visual7w: Grounded question
  answering in images,'' in \emph{CVPR}, 2016.

\end{thebibliography}
}

\end{document}